\documentclass{article} 
\usepackage[preprint]{colm2026_conference}

\usepackage{microtype}
\usepackage{hyperref}
\usepackage{amsmath}
\usepackage{amssymb}
\usepackage[capitalise,noabbrev]{cleveref} 
\usepackage{url}

\usepackage{booktabs}
\usepackage{multirow}
\usepackage{array}
\usepackage{tabularx}
\usepackage{makecell}

\usepackage{listings}

\usepackage{algorithm}
\usepackage{algpseudocode}

\usepackage[T1]{fontenc}
\usepackage{listings}
\lstdefinestyle{regexstyle}{
  basicstyle=\ttfamily\small,
  breaklines=true,
  breakatwhitespace=false,
  columns=fullflexible,
  keepspaces=true,
  showstringspaces=false,
  upquote=true,
  frame=single,
  captionpos=b,
  xleftmargin=0pt,
  xrightmargin=0pt,
  postbreak=\mbox{$\hookrightarrow$\space},
  aboveskip=6pt,
  belowskip=6pt
}

\usepackage{lineno}

\definecolor{darkblue}{rgb}{0, 0, 0.5}
\hypersetup{colorlinks=true, citecolor=darkblue, linkcolor=darkblue, urlcolor=darkblue}

\newcommand{\glr}{GL2R}
\newcommand{\method}{JOLT}

\usepackage{tikz}       

\newcommand{\charsplit}[2]{%
\begin{tikzpicture}[baseline=-0.6ex, x=0.36cm, y=0.36cm]

    \foreach \i/\c in {0/t,1/a,2/b,3/l,4/e} {

        \def\xshift{0}

        \foreach \s in {#1} {
            \ifnum\i>\numexpr\s-1\relax
                \pgfmathparse{\xshift + 0.7}
                \xdef\xshift{\pgfmathresult}
            \fi
        }

        \def\fillcolor{white}

        \foreach \a/\b/\clr in {#2} {
            \ifnum\i<\b
                \ifnum\i>\numexpr\a-1\relax
                    \xdef\fillcolor{\clr}
                \fi
            \fi
        }

        \node[
            draw,
            fill=\fillcolor,
            minimum width=0.36cm,
            minimum height=0.36cm,
            inner sep=0pt,
            font=\ttfamily\scriptsize
        ]
        at ({\i+\xshift},0)
        {\c};
    }

\end{tikzpicture}%
}

\title{Joint Optimization for Greedy Longest-match Tokenization}

\author{
Adhiraj Singh\thanks{These authors contributed equally and are the corresponding authors.
} \quad
Deepanshu Mody\footnotemark[1] \quad
Ghina Al Shdaifat \quad
Hamza Alshamy \\
\textnormal{Center for Data Science, New York University} \\
\textnormal{New York, NY, USA} \\
\texttt{\{as19687,dm6262,gha2009,ha2486\}@nyu.edu} \\
\AND
Adam Wiemerslage \quad
Varshini Reddy \quad
Craig W. Schmidt \\
\textnormal{Kensho Technologies} \\
\textnormal{Cambridge, MA, USA} \\
\texttt{\{adam.wiemerlsage,varshini.bogolu,craig.schmidt\}@kensho.com}
}

\begin{document}

\ifcolmsubmission
\linenumbers
\fi

\maketitle

\begin{abstract}
Recent work shows that subword vocabularies can be trained to directly optimize compression for a specific inference rule, rather than relying on a greedy merge heuristic like Byte Pair Encoding (BPE). For example, ToaST targets split-tree inference and ConvexTok targets shortest-path inference. We extend this approach to greedy left-to-right (\glr{}) longest-match decoding, the fast and widely used inference rule of WordPiece. We introduce \textbf{Joint Optimization for Greedy Longest-match Tokenization (\method{})}, which formulates \glr{} vocabulary learning as an integer program (IP) over vocabulary-selection and segmentation-choice variables. The key ingredient is a set of greedy-consistency constraints that force each pretoken's segmentation to match exactly what \glr{} produces under the selected vocabulary, making the optimized token count equal to the count realized at deployment. To solve the IP at scale, we use a linear program (LP) relaxation that escalates only unresolved pretokens to higher-order segmentations. The relaxation is near-integral: rounded solutions fall within $0.008$--$0.176\%$ of the LP lower bound, certifying near-optimal \glr{} compression on the training scope. The same bound reveals that BPE already sits within $1$--$2\%$ of the LP lower bound on the training scope, confirming the heuristic is near-optimal for \glr{} inference; \method{} closes $89.6$--$99.4\%$ of this gap. On held-out validation across four training scopes ($N \in \{100\text{k},200\text{k},300\text{k},400\text{k}\}$) and two vocabulary sizes ($|V| \in \{32\text{k},64\text{k}\}$), \method{} yields up to $0.78\%$ fewer tokens than BPE, with gains growing with scope. Together, these results show that most of the compression headroom between BPE and the LP lower bound can be recovered through inference-aligned vocabulary optimization.
\end{abstract}

\section{Introduction}

Most NLP tasks use a tokenizer as a first step, converting text into a sequence of tokens drawn from a fixed vocabulary. Byte Pair Encoding \citep[BPE;][]{sennrich2016neural} is among the most frequently used subword tokenizers; WordPiece~\citep{wu2016googles, schuster-nakajima-wordpiece} and UnigramLM~\citep{kudo2018subword} are also widely deployed. Originally proposed as a general-purpose compression algorithm \citep{gage1994new} and later adapted for NLP \citep{sennrich2016neural}, BPE greedily builds a vocabulary by repeatedly merging the most frequent adjacent token pair until it reaches its target size. Producing shorter token sequences is valuable for language models: it increases the amount of text in a given context window, and reduces inference cost. While the effect of compression on downstream accuracy is still contested \citep{galle-2019-investigating, rust-etal-2021-good, ali-etal-2024-tokenizer, goldman-etal-2024-unpacking, schmidt-etal-2024-tokenization}, its practical benefits alone make compression worth optimizing.

Recent work shows that vocabulary training need not be limited to greedy merge heuristics: ToaST \citep{schmidt2026tokenizationsplittrees} and ConvexTok \citep{tempus2026tokenisationconvexrelaxations} show that one can directly optimize compression by formulating vocabulary training as an integer program (IP), relaxing it to a linear program (LP), and extracting a discrete vocabulary via rounding.\footnote{Exact compression-oriented training is NP-complete in several standard formulations~\citep{whittington2025tokenisation,kastreva2025bounded}, so corpus-scale work solves an LP relaxation and rounds to a fixed vocabulary rather than optimizing the full IP.} This re-frames tokenizer design from heuristic, procedure-driven training to an inference-specific optimization problem. There is therefore no single notion of ``best compressing tokenizer'' independent of inference rule; optimality must be stated relative to the inference rule.

Greedy left-to-right (\glr{}) longest-match decoding is a simple and fast tokenization inference rule, most commonly used by WordPiece. Given a vocabulary, \glr{} scans the input left to right and at each position emits the longest vocabulary token matching the bytes starting there, advancing by its length; with a full set of single-byte tokens in the vocabulary, this always succeeds without unknown tokens. \citet{uzan2024greed} show that \glr{} is effective even when applied to vocabularies trained by other methods. However, there has not been a vocabulary construction technique designed specifically to work with \glr{}. We develop \textbf{Joint Optimization for Greedy Longest-match Tokenization (\method{})}, which follows the same high-level recipe as ToaST and ConvexTok (integer program $\rightarrow$ LP relaxation $\rightarrow$ rounding to a fixed vocabulary) but formulates the model specifically for \glr{} inference, making the optimized token count equal to the one realized at deployment.

Empirically, \method{} reduces greedy-L2R token counts on held-out text by up to $0.78\%$ relative to BPE at equal vocabulary size across four training scopes ($N \in \{100\text{k},200\text{k},300\text{k},400\text{k}\}$), with larger scopes consistently yielding equal or better compression. Beyond the compression gains, the LP bound provides a structural diagnostic: BPE's token count on the training scope falls within $1$--$2\%$ of the LP lower bound, revealing that the BPE heuristic is already near-optimal under \glr{} inference. \method{} closes $89.6$--$99.4\%$ of this gap, and its rounded vocabularies fall within $0.008$--$0.176\%$ of the LP bound, so rounding loses almost none of the LP-achievable compression. \method{}'s small margin over BPE on the held-out validation set therefore reflects the limited headroom above the LP bound, not a rounding artifact.

\section{\method{} Model}

\method{} formulates vocabulary learning as a joint integer program over two coupled choices: which candidate tokens enter the vocabulary, and which candidate segmentation of each pretoken is selected. Under \glr{}, the longest matching vocabulary token wins at each position, so a segmentation is \glr{}-realizable only for vocabularies in which no longer token intrudes at any segment boundary. Enforcing this validity condition explicitly (\cref{subsec:final_greedy}) is what couples the vocabulary-selection and segmentation-choice variables. Because the number of candidate segmentations grows exponentially in pretoken length, \method{} does not instantiate them all up front, but initially works with a small number of candidate segmentations per pretoken, adding more as needed.

\subsection{Pretokens, segmentations, and costs}
\label{subsec:final_sets_modeling}

\paragraph{Pretokens.}
We first segment the training corpus into pretokens using a regular expression\footnote{We use a length limited version of the GPT-4o regex. See \cref{app:gpt4oregex16} for more details.} and aggregate identical pretokens by count. Let $P$ be the set of aggregated pretokens in the training data, with frequency $c_p \in \mathbb{N}$ for each pretoken $p \in P$. We work with byte-level tokenization, so we define $n_p = |p|$ as the byte length of pretoken $p$.

For vocabulary training we retain the $N$ highest-count pretokens; throughout the IP below, $P$ denotes this restricted top-$N$ set with counts $c_p$, not the full pretoken vocabulary.

\paragraph{Segmentations.}
A segmentation $s$ partitions a pretoken $p$ into contiguous, non-empty candidate tokens. We depict example segmentations visually divided by $|$, e.g. segmentation $s=\texttt{ta}\,|\,\texttt{ble}$ for pretoken $p=\texttt{table}$. The \emph{order} of $s$ is the number of pieces (tokens) in the partition. An order-$r$ segmentation is determined by choosing $r-1$ internal split positions among the $n_p-1$ byte gaps, so there are exactly $\binom{n_p-1}{r-1}$ order-$r$ candidates for $p$. Summing over orders shows the total grows exponentially in $n_p$:
\begin{equation}
\sum_{r=1}^{n_p}\binom{n_p-1}{r-1}=2^{n_p-1}
\end{equation}
In practice, however, a compression-oriented vocabulary will tokenize most pretokens with a low-order segmentation, so only a small fraction of the $2^{n_p-1}$ candidates need to be explicitly instantiated.

To avoid the exponential number of potential segmentations, we instantiate only segmentations up to a bounded order per pretoken. Let $\kappa_p$ denote the maximum segmentation order currently instantiated for $p$. For a given $\kappa_p$, let $S_p$ be the finite set of segmentation choices for $p$: all $s$ with $\mathrm{order}(s) \le \kappa_p$, together with a fallback $\mathcal{F}$. The fallback is not itself a segmentation of $p$; choosing $\mathcal{F}$ means that none of the currently instantiated segmentations suffice and the solver would prefer a partition of order greater than $\kappa_p$. We initially set $\kappa_p=2$, so $S_p$ contains only the whole pretoken, all order-2 segmentations, and $\mathcal{F}$; these are the only segmentation options for $p$ in the first round. When the solver selects $\mathcal{F}$ for $p$, $\kappa_p$ is incremented and the next order of segmentations is appended to $S_p$ (\cref{subsec:final_escalation}). Thus, 
\begin{equation}
|S_p| = \sum_{r=1}^{\kappa_p}\binom{n_p-1}{r-1} + 1.
\label{eq:rev_Sp_card}
\end{equation}
Escalation then increases $\kappa_p$ and enriches $S_p$ only for those pretokens (\cref{subsec:final_escalation}), rather than instantiating higher-order segmentations for every $p\in P$ up front. This selective escalation avoids instantiating high-order segmentations unnecessarily, reducing LP size and solve time.

\paragraph{Tokens and costs.}
Training proceeds in a sequence of \emph{rounds}. In each round, every pretoken $p$ has a fixed limit $\kappa_p$ on segmentation order, with a fixed finite set $S_p$. We build and solve the IP (or its LP relaxation); pretokens that select the fallback $\mathcal{F}$ trigger an increase in $\kappa_p$ and a rebuild before the next round (\cref{subsec:final_escalation}).

\cref{fig:admissible_splits}(a) gives an example for the pretoken $p=\texttt{table}$ with $n_p=5$. There are $2^{n_p-1}=16$ potential segmentations, but with the initial $\kappa_p=2$ only six are instantiated choices: the whole pretoken, four order-2 segmentations, and $\mathcal{F}$. When a pretoken selects the fallback, its $\kappa_p$ is increased by one in the next round and those new candidates are appended (\cref{subsec:final_escalation}). \cref{fig:admissible_splits}(b) shows the six extra order-3 segmentations added when $\kappa_p$ increases to $3$; all $2^{n_p-1}$ segmentations enter $S_p$ only if $\kappa_p$ reaches $n_p$.

\begin{figure}[t]
\centering
\small
\renewcommand{\arraystretch}{1.0}
\setlength{\tabcolsep}{2pt}

\begin{minipage}[t]{0.48\linewidth}
\centering

\textbf{(a) Round~1}\\
{\scriptsize ($6$ instantiated / $16$ possible)}

\vspace{0.3em}

\resizebox{!}{1.7cm}{%
\begin{tabular}{@{}clc@{}}
\toprule
\textbf{Variable} & \textbf{Visual} & $\mathrm{cost}(p,s)$ \\
\midrule
$z_{\texttt{table},\,\texttt{table}}$
& \charsplit{}{}
& $1$ \\

$z_{\texttt{table},\,\texttt{t}\,|\,\texttt{able}}$
& \charsplit{1}{}
& $2$ \\

$z_{\texttt{table},\,\texttt{ta}\,|\,\texttt{ble}}$
& \charsplit{2}{}
& $2$ \\

$z_{\texttt{table},\,\texttt{tab}\,|\,\texttt{le}}$
& \charsplit{3}{}
& $2$ \\

$z_{\texttt{table},\,\texttt{tabl}\,|\,\texttt{e}}$
& \charsplit{4}{}
& $2$ \\

$z_{\texttt{table},\,\mathcal{F}}$
& {\scriptsize ---}
& $\kappa_p{+}1$ \\
\bottomrule
\end{tabular}
}

\end{minipage}
\hfill
\begin{minipage}[t]{0.48\linewidth}
\centering

\textbf{(b) Added if fallback selected in Round~1}\\
{\scriptsize (six new order-3 segmentations)}

\vspace{0.3em}

\resizebox{!}{1.7cm}{%
\begin{tabular}{@{}clc@{}}
\toprule
\textbf{Variable} & \textbf{Visual} & $\mathrm{cost}(p,s)$ \\
\midrule
$z_{\texttt{table},\,\texttt{t}\,|\,\texttt{a}\,|\,\texttt{ble}}$
& \charsplit{1,2}{}
& $3$ \\

$z_{\texttt{table},\,\texttt{t}\,|\,\texttt{ab}\,|\,\texttt{le}}$
& \charsplit{1,3}{}
& $3$ \\

$z_{\texttt{table},\,\texttt{t}\,|\,\texttt{abl}\,|\,\texttt{e}}$
& \charsplit{1,4}{}
& $3$ \\

$z_{\texttt{table},\,\texttt{ta}\,|\,\texttt{b}\,|\,\texttt{le}}$
& \charsplit{2,3}{}
& $3$ \\

$z_{\texttt{table},\,\texttt{ta}\,|\,\texttt{bl}\,|\,\texttt{e}}$
& \charsplit{2,4}{}
& $3$ \\

$z_{\texttt{table},\,\texttt{tab}\,|\,\texttt{l}\,|\,\texttt{e}}$
& \charsplit{3,4}{}
& $3$ \\
\bottomrule
\end{tabular}
}

\end{minipage}

\vspace{0.2em}


\caption{%
Instantiated segmentations $S_p$ for pretoken $p=\texttt{table}$ ($n_p{=}5$).
\textbf{(a)}~Round~1 ($\kappa_p{=}2$): whole pretoken, four order-2 segmentations, and fallback $\mathcal{F}$.
\textbf{(b)}~Round~2: six order-3 segmentations added when $\mathcal{F}$ was selected in round~1 (\cref{subsec:final_escalation}).
}
\label{fig:admissible_splits}
\end{figure}

Let $B$ denote the set of all 256 single-byte tokens which are always retained in the extracted vocabulary (\cref{eq:rev_byte}) to avoid unknown tokens. For each $p\in P$ and non-fallback $s\in S_p$, let $E_{p,s}$ be the set of distinct token strings (byte subsequences) used by segmentation $s$. The global \emph{candidate token set} for that round is
\begin{equation}
T \;=\; B \;\cup\; \bigcup_{p\in P}\;\bigcup_{s\in S_p \setminus \{\mathcal{F}\}} E_{p,s}.
\label{eq:candidate_T}
\end{equation}
Elements $t\in T$ are byte strings; whenever any $S_p$ grows, new segmentations may introduce token strings absent from the current $T$, so $T$ is \emph{recomputed} before the next LP solve.

For non-fallback segmentations, $\mathrm{cost}(p,s)$ equals the order of $s$ (token count under $s$). The fallback enters the objective through a stage-dependent cost:
\begin{equation}
\mathrm{cost}(p,\mathcal{F}) =
\begin{cases}
\kappa_p + 1, & \text{intermediate rounds},\\
n_p, & \text{final repricing round},
\end{cases}
\label{eq:rev_fb_cost}
\end{equation}
In intermediate rounds, $\mathrm{cost}(p,\mathcal{F})=\kappa_p+1$ is intentionally one token above the largest cost among currently instantiated segmentations of order $\le\kappa_p$, so any non-fallback segmentation is strictly cheaper than $\mathcal{F}$. \cref{subsec:final_escalation} uses this pricing to decide when to enrich $S_p$; the final round sets $\mathrm{cost}(p,\mathcal{F})=n_p$ so the reported score reflects the worst-case \glr{} token count for any remaining fallback pretokens once $S_p$ has converged.

\subsection{Decision variables and objective}
\label{subsec:final_vars_obj}

\paragraph{Vocabulary-selection variables.}
For each candidate token $t \in T$, define
\begin{equation}
x_t \in \{0,1\}
\quad \forall t \in T,
\label{eq:rev_x_domain}
\end{equation}
where $x_t=1$ means token $t$ is in the vocabulary. For example, for pretoken \texttt{table}, $x_{\texttt{tab}}=1$ means \texttt{tab} is globally available as a candidate token.

\paragraph{Segmentation-choice variables.}
For each $p \in P$ and each $s \in S_p$, we introduce a binary decision variable
\begin{equation}
z_{p,s} \in \{0,1\}
\quad \forall p \in P,\ \forall s \in S_p,
\label{eq:rev_z_domain}
\end{equation}
where $z_{p,s}=1$ indicates that segmentation $s$ is selected for pretoken $p$; the greedy-consistency constraints (\cref{subsec:final_greedy}) then enforce that the active segmentation is exactly the one \glr{} produces. Because $\mathrm{cost}(p,s) < \mathrm{cost}(p,\mathcal{F})$ for every non-fallback $s$, selecting $\mathcal{F}$ (i.e., $z_{p,\mathcal{F}}=1$ in the integer problem, or LP mass on $z_{p,\mathcal{F}} \ge \theta$ in the relaxation; we use $\theta=0.01$) is the signal that the current $S_p$ is too coarse; \cref{subsec:final_escalation} acts on that signal to increase $\kappa_p$ and reduce the use of fallbacks.

\paragraph{Optimization objective.}
We minimize the corpus-weighted token count: each pretoken $p$ selects exactly one segmentation $s \in S_p$, paying $\mathrm{cost}(p,s)$ tokens, weighted by the aggregated count $c_p$. The objective is:
\begin{equation}
\min_{x,z} \sum_{p \in P} \sum_{s \in S_p} c_p \,\mathrm{cost}(p,s)\, z_{p,s}
\label{eq:rev_obj}
\end{equation}

\subsection{Feasibility constraints}
\label{subsec:final_feasibility}

\paragraph{Byte-completeness constraint.}
Single-byte tokens must be included in the vocabulary, ensuring any input can be tokenized without unknown-token failures:
\begin{equation}
x_b = 1 \quad \forall b \in B.
\label{eq:rev_byte}
\end{equation}

\paragraph{Vocabulary-budget constraint.}
We optimize a vocabulary of desired size $|V|$:
\begin{equation}
\sum_{t \in T} x_t \le |V|.
\label{eq:rev_budget}
\end{equation}

\paragraph{Exactly-one-segmentation constraint.}
Exactly one segmentation is active per pretoken:
\begin{equation}
\sum_{s \in S_p} z_{p,s} = 1
\quad \forall p \in P.
\label{eq:rev_cover}
\end{equation}

\paragraph{Linking constraint.}
For non-fallback $s$, recall from \cref{subsec:final_sets_modeling} that $E_{p,s}$ is the set of tokens $t$ used in segmentation $s$ of pretoken $p$. A segmentation can be activated only if all of its tokens are selected in the vocabulary:
\begin{equation}
z_{p,s} \le x_t
\quad \forall p \in P,\ \forall s \in S_p \setminus \{\mathcal{F}\},\ \forall t \in E_{p,s}.
\label{eq:rev_link}
\end{equation}
For example, splitting \texttt{table} as \texttt{ta}\,|\,\texttt{ble} requires both \texttt{ta} and \texttt{ble} to be in the vocabulary. The blue rows of \cref{fig:split_constraints} show this for two example segmentations.

\subsection{Greedy-consistency constraints}
\label{subsec:final_greedy}

\begin{figure*}[t]
\centering
\small
\renewcommand{\arraystretch}{1.0}
\setlength{\tabcolsep}{2pt}

\begin{minipage}[t]{0.49\textwidth}
\vspace{0pt}
\centering
\resizebox{\linewidth}{!}{%
\begin{tabular}{@{}llc@{}}
\toprule
\multicolumn{3}{c}{%
  \textbf{(a)} $s=\texttt{ta}\,|\,\texttt{ble}$
} \\
\midrule
\textbf{Type} & \textbf{Inequality} & \textbf{Segmentation} \\
\midrule
\multirow{2}{*}{Linking}
& $z_{\texttt{table},\,\texttt{ta}\,|\,\texttt{ble}} \le x_{\texttt{ta}}$
& \charsplit{2}{0/2/blue!14} \\[2pt]
& $z_{\texttt{table},\,\texttt{ta}\,|\,\texttt{ble}} \le x_{\texttt{ble}}$
& \charsplit{2}{2/5/blue!14} \\
\midrule
\multirow{3}{*}{\shortstack[l]{Greedy\\consist.}}
& $z_{\texttt{table},\,\texttt{ta}\,|\,\texttt{ble}} \le 1 - x_{\texttt{tab}}$
& \charsplit{3}{0/3/red!12} \\[2pt]
& $z_{\texttt{table},\,\texttt{ta}\,|\,\texttt{ble}} \le 1 - x_{\texttt{tabl}}$
& \charsplit{4}{0/4/red!12} \\[2pt]
& $z_{\texttt{table},\,\texttt{ta}\,|\,\texttt{ble}} \le 1 - x_{\texttt{table}}$
& \charsplit{}{0/5/red!12} \\
\bottomrule
\end{tabular}
}
\end{minipage}
\hfill
\begin{minipage}[t]{0.49\textwidth}
\vspace{0pt}
\centering
\resizebox{\linewidth}{!}{%
\begin{tabular}{@{}llc@{}}
\toprule
\multicolumn{3}{c}{%
  \textbf{(b)} $s=\texttt{ta}\,|\,\texttt{b}\,|\,\texttt{le}$
} \\
\midrule
\textbf{Type} & \textbf{Inequality} & \textbf{Segmentation} \\
\midrule
\multirow{3}{*}{Linking}
& $z_{\texttt{table},\,\texttt{ta}\,|\,\texttt{b}\,|\,\texttt{le}} \le x_{\texttt{ta}}$
& \charsplit{2,3}{0/2/blue!14} \\[2pt]
& $z_{\texttt{table},\,\texttt{ta}\,|\,\texttt{b}\,|\,\texttt{le}} \le x_{\texttt{b}}$
& \charsplit{2,3}{2/3/blue!14} \\[2pt]
& $z_{\texttt{table},\,\texttt{ta}\,|\,\texttt{b}\,|\,\texttt{le}} \le x_{\texttt{le}}$
& \charsplit{2,3}{3/5/blue!14} \\
\midrule
\multirow{5}{*}{\shortstack[l]{Greedy\\consist.}}
& $z_{\texttt{table},\,\texttt{ta}\,|\,\texttt{b}\,|\,\texttt{le}} \le 1 - x_{\texttt{tab}}$
& \charsplit{3}{0/3/red!12} \\[2pt]
& $z_{\texttt{table},\,\texttt{ta}\,|\,\texttt{b}\,|\,\texttt{le}} \le 1 - x_{\texttt{tabl}}$
& \charsplit{4}{0/4/red!12} \\[2pt]
& $z_{\texttt{table},\,\texttt{ta}\,|\,\texttt{b}\,|\,\texttt{le}} \le 1 - x_{\texttt{table}}$
& \charsplit{}{0/5/red!12} \\[2pt]
& $z_{\texttt{table},\,\texttt{ta}\,|\,\texttt{b}\,|\,\texttt{le}} \le 1 - x_{\texttt{bl}}$
& \charsplit{2,4}{2/4/red!12} \\[2pt]
& $z_{\texttt{table},\,\texttt{ta}\,|\,\texttt{b}\,|\,\texttt{le}} \le 1 - x_{\texttt{ble}}$
& \charsplit{2}{2/5/red!12} \\
\bottomrule
\end{tabular}
}
\end{minipage}

\caption{%
Linking and greedy-consistency rows for $p=\texttt{table}$.
\textbf{(a)}~$s=\texttt{ta}\,|\,\texttt{ble}$.
\textbf{(b)}~$s=\texttt{ta}\,|\,\texttt{b}\,|\,\texttt{le}$.
Blue: required token; red: forbidden longer prefix at that segment start.
}
\label{fig:split_constraints}
\end{figure*}

The feasibility constraints above (\crefrange{eq:rev_byte}{eq:rev_link}) ensure budget compliance and that an activated segmentation uses tokens selected in the vocabulary, but they do not yet enforce that it is the segmentation \glr{} would actually produce for the given vocabulary.
A segmentation $s$ is \emph{\glr{}-realizable} (under vocabulary $V$) if it is exactly the segmentation \glr{} longest-prefix decoding emits on pretoken $p$.
\emph{Greedy-consistency} means that every choice with $z_{p,s}=1$ is \glr{}-realizable; we enforce this with \cref{eq:rev_block}.

A \emph{segment} $g$ is one contiguous piece of a segmentation $s$; for $s=\texttt{ta}\,|\,\texttt{b}\,|\,\texttt{le}$ the segments are \texttt{ta}, \texttt{b}, and \texttt{le}. Let $G_s^{-}$ denote the \emph{non-final} segments of $s$ (here \texttt{ta} and \texttt{b}); the final segment has nothing downstream to displace, so it imposes no constraint. For byte strings $u,v$ we write $u \preceq v$ if $u$ is a prefix of $v$, and $u \prec v$ if additionally $u \ne v$. Under \glr{}, $s$ is \glr{}-realizable only if at the start of each non-final segment, no longer vocabulary token extends past that segment, otherwise \glr{} would emit the longer token and shift all downstream boundaries. The red rows of \cref{fig:split_constraints} illustrate this for two example segmentations.

Consider a non-final segment $g \in G_s^{-}$, and let $w$ be the suffix of $p$ that starts at $g$ (so $w$ begins with $g$, i.e.\ $g \preceq w$). For example, with $p=\texttt{table}$ and $g=\texttt{ta}$ we have $w=\texttt{table}$; with $g=\texttt{b}$ we have $w=\texttt{ble}$. The \emph{forbidden-prefix set} $L_{p,s,g}$ comprises candidate tokens that strictly extend $g$ while still matching the upcoming bytes of $p$:
\begin{equation}
L_{p,s,g} \;=\; \bigl\{\, t \in T \;:\; g \prec t \preceq w \,\bigr\} \quad \forall\,p \in P,\ \forall\, s \in S_p \setminus \{\mathcal{F}\},\ \forall\, g \in G_s^{-}.
\label{eq:blocking}
\end{equation}

For $p=\texttt{table}$, $g=\texttt{ta}$, and $w=\texttt{table}$, \cref{eq:blocking} yields $L_{p,s,g}=\{\texttt{tab},\texttt{tabl},\texttt{table}\}$. Note $g \prec t$ alone is not enough: \texttt{tax} extends \texttt{ta} but is not in $L_{p,s,g}$, since $\texttt{tax} \not\preceq \texttt{table}$ means \glr{} never considers it here. A forbidden longer prefix need not be a segment of $s$ (e.g.\ $\texttt{tab}\in L_{p,s,g}$ for \texttt{ta}\,|\,\texttt{ble} although \texttt{tab} is not one of its segments).

If any $t \in L_{p,s,g}$ is in the vocabulary, $s$ cannot be selected:
\begin{equation}
z_{p,s} \le 1 - x_t
\qquad \forall\, p \in P,\ \forall\, s \in S_p \setminus \{\mathcal{F}\},\ \forall\, g \in G_s^{-},\ \forall\, t \in L_{p,s,g}.
\label{eq:rev_block}
\end{equation}

Together with the linking constraint \cref{eq:rev_link}, which guarantees each segment string is itself in the vocabulary, \cref{eq:rev_block} ensures that whenever $z_{p,s}=1$ each chosen segment is the \emph{longest} vocabulary match at its start position, exactly the segmentation \glr{} longest-prefix decoding emits with vocabulary $V$.

\subsection{Fallback-driven escalation}
\label{subsec:final_escalation}

\Cref{eq:rev_block} must be enforced for every segmentation variable we instantiate, so the size of the IP in \crefrange{eq:rev_obj}{eq:rev_block} is driven by $|S_p|$ in each round. \cref{subsec:final_sets_modeling} explained why naively instantiating all orders for every pretoken in the top-$N$ set would introduce too many variables and constraints, making the LP slow to solve; here we describe the schedule that keeps each solve tractable while preserving a globally coupled objective.

\paragraph{Fallback as model-size control.}
Every pretoken has a distinct $\mathcal{F}\in S_p$ with variable $z_{p,\mathcal{F}}$; setting it to $1$ pays the higher $\mathrm{cost}(p,\mathcal{F})$ rather than committing to any instantiated segmentation. This is not a soft penalty but a \emph{model-size control}: it lets us omit higher-order segmentations until the solver signals the current family is too coarse for $p$.

\paragraph{Fallback-driven iterative escalation.}
Training solves a sequence of LP relaxations of the globally coupled IP, each with the same structure \crefrange{eq:rev_byte}{eq:rev_block} and objective \cref{eq:rev_obj}, but with pretoken segmentation sets $S_p$ growing from round to round, and the candidate set $T$ recomputed accordingly.

\emph{Round~1.} For every $p\in P$, set $\kappa_p=2$ and initialize $S_p$ to the unsplit pretoken, all order-2 segmentations (one internal split per byte gap), and $\mathcal{F}$. To shrink the first LP, the $N_{\mathrm{top}}{=}8000$ highest-count pretokens start with $\kappa_p{=}1$ (whole pretoken and $\mathcal{F}$ only): these extremely frequent tokens (common words like \texttt{the}, \texttt{is}, \texttt{in}, and common punctuation) carry the bulk of corpus mass and are near-certain vocabulary members as whole units regardless of budget, making their order-2 splits almost never worthwhile; omitting them removes a large fraction of variables and greedy-consistency rows (\cref{eq:rev_block}) with little empirical impact. Any pretoken that nonetheless activates $\mathcal{F}$ is promoted to the full order-2 family before the next round.

\emph{Escalation loop.} After the solution of each LP (\cref{subsec:final_lp_round}) with optimal fractional values $(x^\star,z^\star)$ in round~$i$, read off the fallback-active set $F_i = \{p\in P: z_{p,\mathcal{F}}^\star \ge \theta\}$ from the fractional solution. Only pretokens in $F_i$ have $\kappa_p$ incremented (up to cap $M{=}3$), their order-$(\kappa_p{+}1)$ segmentations appended to $S_p$, $T$ and the full IP rebuilt, and re-solved; \cref{fig:admissible_splits}(b) illustrates the order-3 segmentations added for $p=\texttt{table}\in F_i$. The loop continues until $F_i=\emptyset$ or no $\kappa_p$ changes.

\emph{Final repricing.} When the structure stabilizes, one last solve freezes each $S_p$ and sets $\mathrm{cost}(p,\mathcal{F})=n_p$ via \cref{eq:rev_fb_cost}, so resolved pretokens carry their exact \glr{} inference cost while any remaining fallback mass is assigned a conservative byte-level upper bound. LP relaxation and vocabulary extraction from the final fractional solution are described in the next subsection.

\subsection{LP relaxation and weighted rounding}
\label{subsec:final_lp_round}

The IP instances in \crefrange{eq:rev_obj}{eq:rev_block} are too large to solve to optimality at corpus scale, even after iterative enrichment of the segmentation families.
Following ToaST, \method{} works with an LP relaxation that replaces the binary domains in \crefrange{eq:rev_x_domain}{eq:rev_z_domain} by 
\begin{equation}
0 \le x_t \le 1
\quad \forall t \in T,
\label{eq:rev_x_domain_relaxed}
\end{equation}
\begin{equation}
0 \le z_{p,s} \le 1
\quad \forall p \in P,\ \forall s \in S_p.
\label{eq:rev_z_domain_relaxed}
\end{equation}

Solving the resulting LP with Gurobi~\citep{gurobi} yields optimal fractional values $(x^\star,z^\star)$. Relaxing integrality allows segmentation mass to split fractionally across $S_p$, but the relaxation remains globally coupled across pretokens: high-count $p$ still pulls token mass toward compressing segmentations that survive \glr{} linking and greedy-consistency coupling. In practice the final-round solution is empirically near-integral on our instances (many $x_t^\star$ near $0$ or $1$), which makes a simple rounding rule effective to construct a budget-feasible vocabulary under \cref{eq:rev_budget} without solving the full IP to optimality. The optimal objective value is a valid lower bound on the corresponding instantiated IP objective.

\paragraph{Contribution-based rounding.} We extract the vocabulary from $(x^\star,z^\star)$ using the contribution-based weighted rounding of ToaST. The rule is unchanged: although \method{}'s segmentation variables $z_{p,s}$ range over \glr{}-realizable segmentations rather than split-tree ones, rounding treats them only as generic segmentation variables, so the procedure transfers verbatim. Tokens with $x_t^\star \ge 1-\epsilon$ are admitted to $V$ outright (\cref{app:rounding}); the remaining budget is filled from the fractional tier by \emph{objective contribution} $C^\star_t$, the count-weighted $z^\star$-mass flowing through $t$ (\cref{eq:cstar} in \cref{app:rounding}). Selection therefore favors tokens the relaxation already used to compress high-frequency pretokens, rather than tokens with large $x_t^\star$ but little segmentation support. Inference applies \glr{} decoding with $V$.

\section{Experimental Setup}
\label{sec:exp_setup}

\subsection{Data and pretokenization}
\label{subsec:data_universes_v2}

We train and evaluate all tokenizers on MiniPile~\citep{kaddour2023minipilechallengedataefficientlanguage}, a subset of The Pile~\citep{gao2020pile}, using a document-level 75\%/25\% train/validation split.
Both splits are pretokenized with \texttt{GPT4O\_REGEX\_16} (details in \cref{app:gpt4oregex16}), which segments text into words, numbers, punctuation, and whitespace runs; tokens cannot cross pretoken boundaries.
We write $P_{\mathrm{val}}$ for the full set of aggregated pretoken types on the validation holdout (100\% of validation count mass); all token counts, \% decrease, and R\'enyi efficiency in \cref{sec:results} are evaluated on~$P_{\mathrm{val}}$.

\subsection{Baselines and Configurations}
\label{subsec:scope_ablation_v2}

\paragraph{Baselines.}
BPE, WordPiece, and UnigramLM are trained on the full training corpus using standard HuggingFace pipelines.
\textit{BPE-native} decodes with the standard BPE merge rule; \textit{BPE-greedy} applies \glr{} longest-prefix decoding to the same vocabulary, isolating the effect of inference rule from vocabulary construction.
WordPiece-greedy and UnigramLM-greedy analogously pair standard vocabularies with \glr{} decoding at evaluation.
UnigramLM-greedy serves as a diagnostic for training-objective mismatch: it applies the same \glr{} inference as \method{} but was trained to maximize unigram language-model likelihood rather than minimize token count, so its substantially higher token counts reflect training-objective divergence.
All methods are evaluated at matched vocabulary sizes $|V| \in \{32{,}000, 64{,}000\}$; \% improvements are relative to BPE-native merge decoding.

\paragraph{\method{} configurations.}
We evaluate \method{} at four training scopes ($N \in \{100\text{k}, 200\text{k}, 300\text{k}, 400\text{k}\}$) and both vocabulary budgets, with all runs capped at escalation order $M{=}3$.\footnote{We verified $M{=}3$ is sufficient: $M{=}4$ at $|V|{=}32$k/$N{=}100$k emits $368.7$M tokens on $P_{\mathrm{val}}$ vs.\ $368.6$M at $M{=}3$ (${\approx}0.05\%$ apart). Higher order adds substantial LP size with no compression gain.}
Because corpus-scale LP solves are far more expensive than standard baseline training, \method{} trains exclusively on the top-$N$ pretokens (the set~$P$ of \cref{subsec:final_sets_modeling}), while baselines train on the full distribution; both are evaluated on the same~$P_{\mathrm{val}}$.
Pretoken frequencies are highly skewed, so even the smallest scope ($N{=}100$k) covers $\approx 95.8\%$ of training count mass, rising to $\approx 98.3\%$ at $N{=}400$k; nevertheless, this asymmetry can disadvantage \method{} for tail pretokens not seen during optimization, so reported validation margins may understate its compression quality on the training scope where LP certificates apply.
The two larger scopes ($N{=}300$k and $N{=}400$k) use the round-2-only protocol of \cref{subsec:extended_scope} to assess training scope sensitivity.

\subsection{Extended-scope protocol ($N{=}300$k and $N{=}400$k)}
\label{subsec:extended_scope}

Full-pipeline runs at $N{=}100$k/$200$k provide our LP optimality certificates; extended scopes at $N{=}300$k/$400$k use a round-2-only protocol within our compute budget and are reported to assess whether validation compression continues to improve as $N$ grows, not to claim an equivalent final-round bound.
At $N{=}300$k and $N{=}400$k the final repricing LP becomes prohibitively expensive within our compute budget (See the full breakdown of solve times in \cref{tab:lp_solve_diag}, \cref{app:compute_time}), so we stop after iter-2 and skip the final LP; the iter-2 vocabulary from contribution-based rounding is directly used for validation scoring in \cref{tab:main_results}.

To keep gap-closed and round-gap diagnostics comparable across all training scopes in \cref{tab:lp_round_diag}, we report the raw Gurobi iter-2 LP objective and the greedy-L2R token count on~$P$ from the iter-2 vocabulary.
We do not correct the iter-2 fallback cost ($\kappa_p{+}1{=}3$) to the final-round byte-level penalty $n_p$: doing so after the fact would invalidate the relaxation bound, since $z^\star$ was optimized under the original costs (\cref{app:extended_scope}).

Because the full pipeline remains affordable at $N{=}100$k and $N{=}200$k, we use those runs as a sanity check: extracting the iter-2 vocabulary and rounded count from saved checkpoints recovers full-pipeline rounded objectives to within $0.03\%$ and validation token counts to within ${\approx}0.5$M tokens (\cref{app:extended_scope}), justifying the round-2-only protocol at larger~$N$.

\section{Results and Analysis}
\label{sec:results}

\subsection{Optimality Bounds and Gap Closure}
\label{subsec:solver_bounds}

\begin{figure}[t]
\centering
\includegraphics[width=0.72\columnwidth]{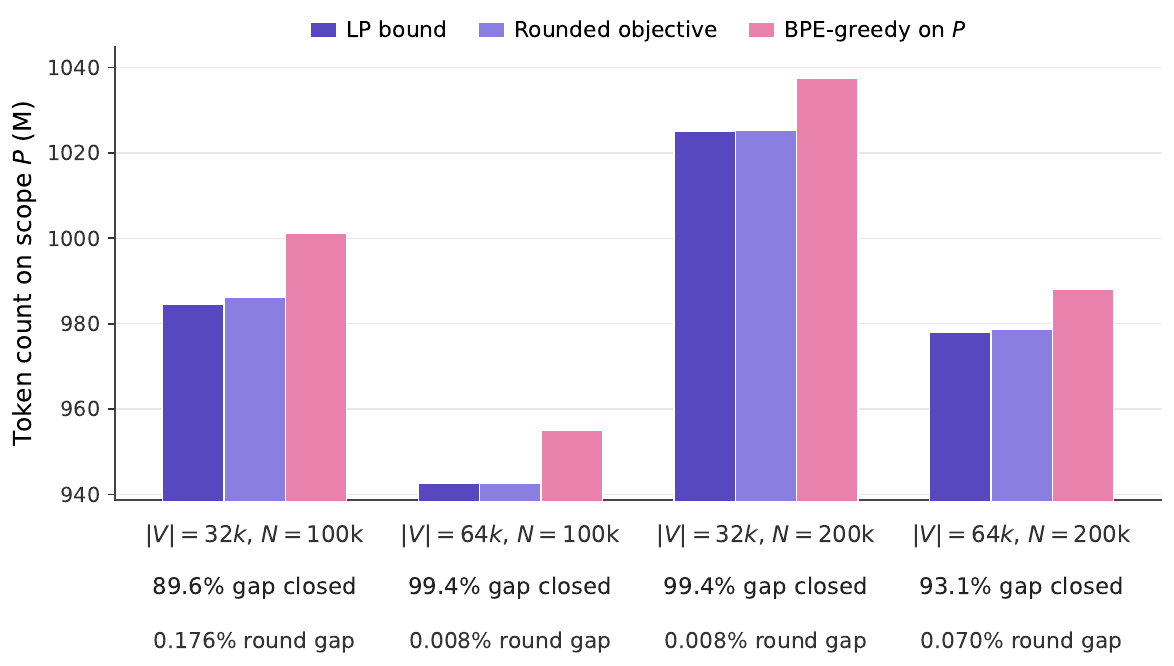}
\caption{LP bound, rounded greedy-L2R objective, and BPE-greedy token count on training scope~$P$ (in millions), with gap-closed \% and round-gap annotated per group. \method{} closes $89.6$--$99.4\%$ of the BPE-greedy-to-LP slack; rounded solutions sit within $0.008$--$0.176\%$ of the LP bound. Full-pipeline runs only ($N \in \{100\text{k},200\text{k}\}$); extended-scope rows in \cref{tab:lp_round_diag}.}
\label{fig:convergence_gap}
\end{figure}

\Cref{fig:convergence_gap} shows the LP bound, rounded \method{} token count, and BPE-greedy on the training scope~$P$ (not on~$P_{\mathrm{val}}$) for the four full-pipeline runs ($N \in \{100\text{k},200\text{k}\}$, both vocabulary budgets).
The LP objective is a \emph{valid lower bound} on achievable \glr{} token count under our formulation, and all four instances solve to \texttt{OPTIMAL} (\cref{app:lp_diag,app:gurobi_setup}).
The rounded score exceeds this LP bound by only $0.008$--$0.176\%$, so the extracted vocabularies nearly attain the relaxation optimum on~$P$.
Because rounded \method{} vocabularies sit this close to the LP bound, any margin over BPE-native on~$P_{\mathrm{val}}$ reflects genuine LP-level compression gains, not rounding loss.
On the matched training scope~$P$ (not on~$P_{\mathrm{val}}$, where tail pretokens outside~$P$ dilute the comparison), \method{} closes $89.6$--$99.4\%$ of the BPE-greedy-to-LP gap, 
confirming most achievable compression under our formulation is recovered.
BPE-greedy already sits near the LP lower bound on~$P$, which is why heuristic training leaves only a small absolute margin on~$P_{\mathrm{val}}$.
Intuitively, BPE's left-to-right merge order naturally favors longer, more frequent substrings, which aligns structurally with \glr{}'s longest-match rule. Even though BPE was not designed for \glr{}, the two procedures share a preference for greedy length maximization that keeps BPE close to the \glr{}-optimal bound.

\subsection{Validation Compression Across Training Scopes}
\label{subsec:main_comparison}

\Cref{tab:main_results} reports greedy-L2R intrinsic metrics on~$P_{\mathrm{val}}$ for all four training scopes.
Rows marked~$^\dagger$ use the round-2-only protocol (\cref{subsec:extended_scope}) and report validation compression only; LP optimality certificates come from the full pipeline at $N{=}100$k/$200$k (\cref{subsec:solver_bounds}).

\begin{table}[t]
\centering
\small
\setlength{\tabcolsep}{2pt}
\renewcommand{\arraystretch}{1.1}

\begin{tabular*}{\linewidth}{@{\extracolsep{\fill}}lcccccc}
\toprule
\multirow{2}{*}{Method} &
\multicolumn{2}{c}{\begin{tabular}[c]{@{}c@{}}Greedy-L2R\\tokens (M)\end{tabular}} &
\multicolumn{2}{c}{\begin{tabular}[c]{@{}c@{}}\% decrease\\vs BPE-native\end{tabular}} &
\multicolumn{2}{c}{\begin{tabular}[c]{@{}c@{}}R\'enyi efficiency\\\(\alpha=2.5\)\end{tabular}} \\
\cmidrule(lr){2-3} \cmidrule(lr){4-5} \cmidrule(lr){6-7}
& 32k & 64k & 32k & 64k & 32k & 64k \\
\midrule

BPE-native merge decoding
& 370.8 & 351.9
& --- & ---
& 0.486 & 0.448 \\

BPE-greedy
& 370.4 & 351.6
& +0.08 & +0.08
& 0.486 & 0.449 \\

WordPiece-greedy
& 372.3 & 352.8
& \(-0.42\) & \(-0.25\)
& 0.487 & 0.449 \\

UnigramLM-greedy
& 476.0 & 460.6
& \(-28.37\) & \(-30.91\)
& 0.352 & 0.334 \\

\midrule

\method{} $N{=}100k$ (ours)
& 368.6 & 354.1
& +0.59 & \(-0.63\)
& 0.485 & 0.450 \\

\method{} $N{=}200k$ (ours)
& 367.9 & 351.1
& +0.77 & +0.23
& 0.485 & 0.448 \\

\method{} $N{=}300k$ (ours)$^\dagger$
& 367.9 & 351.0
& +0.76 & +0.26
& 0.485 & 0.449 \\

\method{} $N{=}400k$ (ours)$^\dagger$
& 367.9 & 350.8
& +0.78 & +0.31
& 0.485 & 0.448 \\

\bottomrule
\end{tabular*}

\caption{%
Intrinsic token-count comparison on~$P_{\mathrm{val}}$ (\cref{subsec:data_universes_v2}). \method{} trains on the top-$N$ pretokens; baselines train on the full corpus (\cref{sec:exp_setup}).
Token counts in millions (M); \% decrease relative to BPE-native (positive = fewer tokens).
$^\dagger$Round-2-only protocol (\cref{subsec:extended_scope}).
Rényi efficiencies cluster tightly across methods; differences in compression appear in token counts, not the Rényi spread.
}
\label{tab:main_results}
\end{table}

At $|V|{=}32$k, \method{} beats BPE-native at every evaluated scope: $+0.59\%$ at $N{=}100$k, $+0.77\%$ at $N{=}200$k, $+0.76\%$ at $N{=}300$k, and $+0.78\%$ at $N{=}400$k.
At $|V|{=}64$k, whether \method{} beats BPE-native depends entirely on scope: $N{=}100$k trails by $0.63\%$, while $N \ge 200$k consistently beats it with the margin growing from $+0.23\%$ to $+0.31\%$ as $N$ increases.
BPE-native and BPE-greedy differ by only $0.08\%$ at both vocabulary budgets, confirming that for BPE vocabularies the inference rule is almost irrelevant to compression. The gap between BPE and \method{} therefore reflects vocabulary construction, not inference-rule mismatch.
WordPiece-greedy, despite being designed for \glr{} inference, attains worse compression than BPE-native ($-0.42\%$ at $32$k, $-0.25\%$ at $64$k), consistent with its training objective---maximizing unigram language-model likelihood rather than minimizing token count---diverging from compression minimization.
The rate of improvement with $N$ differs by vocabulary budget: at $|V|{=}32$k gains plateau near $N{=}200$k but show a slight further increase at $N{=}400$k ($+0.78\%$), while at $|V|{=}64$k compression improves more steadily through $N{=}400$k ($+0.31\%$). The smaller budget saturates earlier because $32$k tokens already cover a large fraction of the high-frequency pretoken tail at $N{=}200$k, whereas the larger vocabulary has more slots to account for the tail of lower frequency pretokens.
Fallback mass after round~1 stays below $1.6\%$ of training count, halves after order-3 escalation, and validation counts improve monotonically with each round; full diagnostics are in \cref{app:fallback_diag}.
In \cref{app:token_zipf,app:token_length_hist} we compare rank-frequency profiles and token-length distributions across tokenizers; the overall profiles are largely similar, with the main structural signature being a modest shift toward length-1 tokens under \method{}.

\section{Conclusion}

We developed a vocabulary construction method tailored to \glr{} longest-match decoding, a widely deployed inference rule that lacked an inference-specific training method. \method{} fills this gap with a joint integer program whose greedy-consistency constraints force each pretoken's segmentation to match exactly what \glr{} realizes under the chosen vocabulary.

The optimizer behaves as designed: rounded \method{} vocabularies close $89.6$--$99.4\%$ of the BPE-greedy-to-LP gap on full-pipeline runs. Yet the headroom is small: BPE already sits within $1$--$2\%$ of the bound, because its left-to-right merge order rewards the long, frequent substrings that \glr{} also prefers at decode time. The residual gap is driven primarily by vocabulary construction rather than the inference rule. Thus, \method{}'s margin --- up to $0.78\%$ fewer held-out tokens, increasing with scope --- comes from reshaping the vocabulary, not from changing inference.

The lasting contribution is therefore not the size of the margin but the ability to certify it relative to an LP lower bound. The LP bound separates optimizer quality from the headroom the inference rule allows, turning ``how much better can a \glr{} tokenizer be?'' from an open question into a measured one. It also provides a reusable test for whether an inference-specific optimizer is worth building for any other decoding rule.

\section*{Limitations}

All experiments use MiniPile, a predominantly English corpus, with a single train/validation split and a fixed regex pretokenizer. Optimality claims are relative to the pretoken boundaries induced by this pretokenizer, and we do not evaluate transfer to languages without whitespace-delimited words. The IP covers only the top-$N$ pretokens ($\approx 95.8$--$98.3\%$ of count-weighted training mass); the tail is excluded entirely from optimization. Within the training scope, segmentations above order cap $M{=}3$ (\cref{subsec:scope_ablation_v2}) are priced conservatively with $n_p$ rather than optimized. Our certificates are empirical LP-gap measurements on the training scope, not proven general approximation bounds: the IP is never solved to integral optimality, and the rounding step carries no worst-case guarantee. Finally, our evaluation is intrinsic only. Our contribution lies in the joint modeling of vocabulary construction and \glr{} segmentation, alongside the certification methodology, rather than a claim of downstream performance.

\bibliography{colm2026_conference}
\bibliographystyle{colm2026_conference}

\appendix
\crefalias{section}{appendix}
\crefalias{subsection}{appendix}
\Crefname{appendix}{Appendix}{Appendices}
\crefname{appendix}{appendix}{appendices}

\section{\method{} Training Algorithm}
\label{app:rev_pseudocode}
The pseudocode in Algorithm~\ref{alg:rev_train} summarizes the full training pipeline and experimental protocol.

\begin{figure*}[t]
\centering
\begin{minipage}{0.95\textwidth}
\begin{algorithm}[H]
  \caption{Greedy-L2R Joint MIP with Fallback-Driven Escalation}
  \label{alg:rev_train}
  \begin{algorithmic}[1] 
  \Require Aggregated pretokens \(P\) with counts \(c_p\), target vocabulary cardinality \(|V|\), byte set
  \(B\), max segmentation order \(M\), fallback-active threshold \(\theta\) (default 0.01)
  \Ensure Learned vocabulary \(V \subseteq T\) with \(|V|\) tokens
  \State Initialize escalation map \(\kappa_p \gets 2\) for all \(p \in P\)
  \State Initialize fallback-active set \(F \gets \varnothing\)
  \Repeat
      \State \(T \gets B\) \Comment{global candidate tokens for \(x_t\)}
      \For{each pretoken \(p \in P\)}
          \If{whole-only shortcut active for \(p\) and \(p \notin F\)}
              \State \(S_p \gets \{\texttt{whole}, \mathcal{F}\}\) \Comment{model-size
  shortcut for top-\(N_{\text{top}}\)}
          \Else
              \State \(S_p \gets \{\texttt{whole}\} \cup
  \bigcup_{r=2}^{\kappa_p}\{\text{order-}r\text{ segmentations of }p\} \cup \{\mathcal{F}\}\)
          \EndIf
          \State Add all tokens used by \(S_p\) to the candidate token set \(T\)
      \EndFor
      \State Create variables \(x_t \in \{0,1\}\ \forall t \in T\), \(z_{p,s} \in \{0,1\}\ 
  \forall p \in P,\ s \in S_p\)
      \State Add constraints:
      \State \quad \eqref{eq:rev_byte}: \(x_b = 1,\ \forall b \in B\) \Comment{byte completeness}
      \State \quad \eqref{eq:rev_budget}: \(\sum_{t \in T} x_t \le |V|\) \Comment{vocabulary budget}
      \State \quad \eqref{eq:rev_cover}: \(\sum_{s \in S_p} z_{p,s}=1,\ \forall p \in P\) \Comment{exactly one
  segmentation per pretoken}
      \State \quad \eqref{eq:rev_link}: \(z_{p,s}\le x_t,\ \forall p,\ \forall s\in S_p\setminus\{\mathcal{F}\},\ \forall t \in
  E_{p,s}\) \Comment{linking}
      \State \quad \eqref{eq:rev_block}: $z_{p,s}\le 1 - x_t,\ \forall p,\ \forall s\in S_p\setminus\{\mathcal{F}\}, \forall g \in G_s^{-},\ \forall t \in L_{p,s,g}$ \Comment{greedy-consistency}
      \State Set objective \eqref{eq:rev_obj}: \(\min \sum_{p \in P}\sum_{s \in S_p}
  c_p\,\mathrm{cost}(p,s)\,z_{p,s}\)
      \State Solve LP relaxation (or MIP variant) with Gurobi
      \State Extract fallback-active set \(F \gets \{p \in P:\ z_{p,\mathcal{F}}^\star \ge
  \theta\}\) \Comment{from \(z^\star\) before rounding}
      \State \(V \gets \textsc{WeightedRound}(x^\star, z^\star, |V|)\) \Comment{per-iteration
   rounded vocabulary}
      \For{each \(p \in F\)}
          \State Escalate only fallback-active pretokens: \(\kappa_p \gets \min(\kappa_p+1,
  M)\)
      \EndFor
  \Until{\(F=\varnothing\) or no \(\kappa_p\) changes}
  \State \emph{Final repricing round:} set \(\mathrm{cost}(p,\mathcal{F}) \gets |p|\), rebuild
   and re-solve LP once
  \State \(V \gets \textsc{WeightedRound}(x^\star, z^\star, |V|)\) \Comment{final vocabulary}
  \State \Return selected vocabulary \(V\), objectives, bounds, fallback stats
  \Statex
  \Procedure{WeightedRound}{$x^\star,\, z^\star,\, |V|$} \Comment{with $\epsilon = 10^{-6}$}
      \State $T_1 \gets \{t \in T :\ x_t^\star \ge 1 - \epsilon\}$
  \Comment{near-integral, accept all; $B \subseteq T_1$ via \eqref{eq:rev_byte}}
      \State $T_F \gets \{t \in T :\ \epsilon < x_t^\star < 1 - \epsilon\}$
  \Comment{fractional candidates}
      \State $\beta \gets |V| - |T_1|$ \Comment{remaining budget from $T_F$}
      \State Compute $C^\star_t \gets \sum_{(p,s):\,t\in E_{p,s}} c_p\, z^\star_{p,s}$ for
  each $t \in T_F$
      \State \Return $T_1 \cup \mathrm{top}\text{-}\beta\,(T_F,\, C^\star_\cdot)$
  \Comment{cardinality $|V|$ by construction}
  \EndProcedure
  \end{algorithmic}
  \end{algorithm}
\end{minipage}
\end{figure*}

\section{Contribution-Based Weighted Rounding}
\label{app:rounding}
We give the full rounding procedure summarized in \cref{subsec:final_lp_round}; it is ToaST's rule applied to \method{}'s LP solution $(x^\star,z^\star)$. Let $V\subseteq T$ denote the vocabulary with desired size $|V|$. Partition $T$ by the LP value $x_t^\star$:
\begin{align*}
  T_1 &= \{t \in T : x_t^\star \ge 1-\epsilon\}, \\
  T_F &= \{t \in T : \epsilon < x_t^\star < 1-\epsilon\}, \\
  T_0 &= \{t \in T : x_t^\star \le \epsilon\},
\end{align*}
with $\epsilon=10^{-6}$. Every $t\in T_1$ is admitted into $V$ outright. The budget still requires $\beta=|V|-|T_1|$ additional tokens, all drawn from the fractional tier $T_F$ (tokens in $T_0$ are discarded).

For each fractional $t\in T_F$, define the \emph{objective contribution}
\begin{equation}
  C^\star_t \;=\; \sum_{p \in P}\;\sum_{s \in S_p \setminus \{\mathcal{F}\} \,:\, t \in E_{p,s}} c_p\, z^\star_{p,s}.
  \label{eq:cstar}
\end{equation}
$C^\star_t$ is the count-weighted $z$-mass for $t$ at the LP optimum: it measures how much of the relaxed objective is already ``buying'' $t$ via the segmentations that depend on it. Rank $T_F$ in descending order of $C^\star_t$ and take the top $\beta$ tokens:
\begin{equation*}
  V \;=\;  T_1 \;\cup\; \textrm{top-}\beta\,(T_F,\, C^\star_\cdot).
\end{equation*}
By construction, exactly $|V|$ tokens are selected.

\section{\texttt{GPT4O\_REGEX\_16} Pretokenization Pattern}
\label{app:gpt4oregex16}

We use \texttt{GPT4O\_REGEX\_16} as the fixed regex pretokenizer for all experiments. This pattern is based on the publicly available GPT-4o/\texttt{o200k\_base}-style \texttt{tiktoken} regex pretokenizer, but with explicit length caps added for computational tractability. The regex is used only to create the initial pretoken/count universe; the learned vocabulary is subsequently optimized by the LP/MIP formulation described in the main text.

The pattern consists of seven alternatives: lower-case-dominant Unicode word spans with optional contractions, upper-case-dominant Unicode word spans with optional contractions, numeric chunks of length 1--3, punctuation/symbol runs, newline-associated whitespace, trailing whitespace, and general whitespace. The bounded version caps word-like spans through 32-character letter/mark groups and caps punctuation or whitespace runs at 16 characters.

\begin{lstlisting}[
style=regexstyle,
caption={Bounded GPT-4o-style regex pretokenizer used in our experiments.},
label={lst:gpt4o_regex16}
]
GPT4O_REGEX_PARTS_16 = [
    r"[^\r\n\p{L}\p{N}]?"
    r"[\p{Lu}\p{Lt}\p{Lm}\p{Lo}\p{M}]{0,32}"
    r"[\p{Ll}\p{Lm}\p{Lo}\p{M}]{1,32}"
    r"(?i:'s|'t|'re|'ve|'m|'ll|'d)?",

    r"[^\r\n\p{L}\p{N}]?"
    r"[\p{Lu}\p{Lt}\p{Lm}\p{Lo}\p{M}]{1,32}"
    r"[\p{Ll}\p{Lm}\p{Lo}\p{M}]{0,32}"
    r"(?i:'s|'t|'re|'ve|'m|'ll|'d)?",

    r"\p{N}{1,3}",
    r" ?[^\s\p{L}\p{N}]{1,16}[\r\n/]{0,16}",
    r"\s{0,15}[\r\n]{1,16}",
    r"\s{1,16}(?!\S)",
    r"\s{1,16}",
]

GPT4O_REGEX_16 = "|".join(GPT4O_REGEX_PARTS_16)
\end{lstlisting}

In this work, \texttt{GPT4O\_REGEX\_16} should therefore be interpreted as a deterministic preprocessing rule, not as a tokenizer training objective or learned vocabulary. Its role is to define stable pretoken boundaries before candidate segmentation generation, fallback escalation, and greedy-L2R vocabulary optimization.

\section{Gurobi Solver Setup and Reproducibility}
\label{app:gurobi_setup}
All LP relaxations and MIP solves are run on a single host with Gurobi 11.0 ~\citep{gurobi} (academic license), Python~3.11, and the official \texttt{gurobipy} Python bindings. We give the exact solver configuration we used so that the experiments can be reproduced from the pretoken/count universe alone.

\paragraph{Hardware.} Runs are executed on single shared-memory nodes (one solve per node) with at least 220~GB of RAM. We do not cap Gurobi's thread count, so it uses all available cores, between 18 and 30 threads across our runs; we do not pin threads or NUMA nodes. No GPU is used. The barrier factorization dominates memory: it peaks at roughly 10~GB for the largest $N{=}100k$ LP (32k, iter-2/final) and roughly 30~GB for the largest $N{=}200k$ LP (32k, iter-2/final), comfortably within the 220--230~GB memory limit we impose.

\paragraph{Solver parameters.} We solve the LP relaxation with Gurobi's default LP method (concurrent, which on our models selects barrier followed by crossover) and default presolve and cut settings; we do not override \texttt{Method}, \texttt{Crossover}, \texttt{Presolve}, or \texttt{Cuts}. We set \texttt{MIPGap=0} and \texttt{ConcurrentMIP=1}. For the final repricing round we rebuild the model with the updated fallback cost ($\mathrm{cost}(p,\mathcal{F}){=}n_p$), accept LP warm-start hints from the previous round (\texttt{LPWarmStart=2}), and raise \texttt{NumericFocus=2} for tighter numerics; the reported solve times in \cref{tab:lp_solve_diag} therefore include this warm start.

\paragraph{Limits.} We impose a per-solve soft memory limit (\texttt{MemLimit}) of 220--230~GB and a generous wall-clock limit (up to $4\times10^{5}$~s for the largest $N{=}200k$ configuration). Neither limit is binding: every solve in \cref{tab:lp_solve_diag} reaches \texttt{OPTIMAL} via barrier, with factorization memory peaking near 30~GB. Solve time, however, spans three orders of magnitude across configurations, from 420~s ($N{=}100k$, 32k, iter-1) to 337{,}686.7~s (${\approx}93.8$~h; $N{=}200k$, 32k, final repricing round), so the $N{=}200k$ runs require multi-day budgets while the $N{=}100k$ runs finish within hours (largest 32k final, 35{,}020.9~s). We log Gurobi's final status, objective bound, simplex/barrier iteration counts, and final residuals after every solve.

\paragraph{Determinism.} We do not set a solver seed and leave Gurobi's concurrent LP at its default, so the LP objective is reproducible only up to floating-point summation order and solver concurrency (these depend on host and thread count). Given the LP solution, however, the rest of the pipeline is fully deterministic: contribution-based rounding breaks ties in \(C^\star_t\) lexicographically on \(t\), so the extracted vocabulary is uniquely determined by the (fixed) pretoken/count input and segmentation-family construction order.

\section{Compute Time}
\label{app:compute_time}

\begin{table}[t]
\centering
\small
\setlength{\tabcolsep}{2pt}
\renewcommand{\arraystretch}{1.08}

\begin{tabular*}{\linewidth}{@{\extracolsep{\fill}}lrrrrrrr}
\toprule
& \multicolumn{2}{c}{Variables} & \multicolumn{3}{c}{Constraints} & & \\
\cmidrule(lr){2-3} \cmidrule(lr){4-6}
Config & \#\(x_t\) & \#\((z_{p,s}{+}z_{p,\mathcal{F}})\) & \#linking & \#consist. & \#other & NumNZs & Solve time \\
\midrule
32k, 100k, i-1 & 372{,}010 & 758{,}732 & 1{,}217{,}770 & 2{,}415{,}746 & 100{,}155 & 8{,}397{,}774 & 420.2\,s \\
32k, 100k, i-2 & 491{,}350 & 1{,}370{,}251 & 3{,}052{,}335 & 9{,}215{,}293 & 100{,}155 & 26{,}396{,}857 & 5{,}123.5\,s \\
32k, 100k, final & 491{,}350 & 1{,}370{,}251 & 3{,}052{,}335 & 9{,}215{,}293 & 100{,}155 & 26{,}396{,}857 & 35{,}020.9\,s \\
64k, 100k, i-1 & 372{,}010 & 758{,}732 & 1{,}217{,}770 & 2{,}415{,}746 & 100{,}155 & 8{,}397{,}774 & 727.6\,s \\
64k, 100k, i-2 & 412{,}236 & 896{,}648 & 1{,}631{,}518 & 3{,}946{,}349 & 100{,}155 & 12{,}464{,}618 & 2{,}039.3\,s \\
64k, 100k, final & 412{,}236 & 896{,}648 & 1{,}631{,}518 & 3{,}946{,}349 & 100{,}155 & 12{,}464{,}618 & 6{,}287.7\,s \\
32k, 200k, i-1 & 722{,}676 & 1{,}586{,}135 & 2{,}572{,}579 & 5{,}169{,}227 & 200{,}155 & 17{,}792{,}423 & 3{,}039.1\,s \\
32k, 200k, i-2 & 1{,}009{,}710 & 3{,}530{,}855 & 8{,}406{,}747 & 26{,}816{,}605 & 200{,}155 & 74{,}987{,}269 & 64{,}795.3\,s \\
32k, 200k, final & 1{,}009{,}710 & 3{,}530{,}855 & 8{,}406{,}747 & 26{,}816{,}605 & 200{,}155 & 74{,}987{,}269 & 337{,}686.7\,s \\
64k, 200k, i-1 & 722{,}676 & 1{,}586{,}135 & 2{,}572{,}579 & 5{,}169{,}227 & 200{,}155 & 17{,}792{,}423 & 2{,}591.9\,s \\
64k, 200k, i-2 & 938{,}098 & 2{,}655{,}184 & 5{,}779{,}726 & 17{,}407{,}409 & 200{,}155 & 49{,}967{,}552 & 22{,}531.3\,s \\
64k, 200k, final & 938{,}098 & 2{,}655{,}184 & 5{,}779{,}726 & 17{,}407{,}409 & 200{,}155 & 49{,}967{,}552 & 69{,}055.7\,s \\
\bottomrule
\end{tabular*}

\caption{LP problem-size and solve-time diagnostics for \method{} ($N{=}100k$ and $N{=}200k$ training scopes).
\#\(x_t\): vocabulary-selection variables; \#\((z_{p,s}{+}z_{p,\mathcal{F}})\): segmentation plus fallback variables; \#consist.: greedy-consistency rows (\cref{eq:rev_block}).
In the configuration column, \(i\) denotes escalation iteration.
}
\label{tab:lp_solve_diag}
\end{table}

Detailed computational results are given in \cref{tab:lp_solve_diag}. The $N{=}200k$ full-pipeline configuration (full $2\!\to\!3$ escalation and a final repricing round) is the most expensive among the full-pipeline runs: end-to-end about \(113\) hours of wall-clock time at the 32k vocabulary budget and about \(26\) hours at 64k, with the 32k final-round LP alone taking 337{,}686.7~s (${\approx}93.8$~h). The wall-clock cost is dominated by the final-round LP solve in \cref{tab:lp_solve_diag}; the earlier rounds and the entire model-build phase together contribute roughly one to two additional hours. The cheaper $N{=}100k$ variant runs in approximately 11 hours at 32k and under 3 hours at 64k (32k final-round LP solve of 35{,}020.9~s), a ${\sim}10\times$ speedup for only a ${\sim}2.9\times$ smaller constraint matrix. Within each scope, cost also varies with how binding the budget is: $|V|{=}64$k is cheaper than $|V|{=}32$k because the LP is less fractional.

\section{Fallback and Escalation Diagnostics}
\label{app:fallback_diag}

This section reports stage-wise fallback diagnostics for all \method{} configurations (\cref{subsec:final_escalation}). Each row gives the number of pretokens with LP mass on $\mathcal{F}$ at or above threshold $\theta{=}0.01$, the count-weighted fallback mass, and greedy-L2R token totals on the training scope~$P$ and validation holdout~$P_{\mathrm{val}}$ after contribution-based rounding at that stage. We report both fallback count and weighted mass by stage because they can move differently under repricing and rounding.

\begin{table}[t]
\centering
\small
\setlength{\tabcolsep}{4pt}
\renewcommand{\arraystretch}{1.15}

\begin{tabular*}{\linewidth}{@{\extracolsep{\fill}}llrrrrr}
\toprule
\multirow{2}{*}{Config} & \multirow{2}{*}{Stage}
& \multicolumn{3}{c}{Fallback}
& \multicolumn{1}{c}{\multirow{2}{*}{\makecell{Train tokens\\on $P$}}}
& \multicolumn{1}{c}{\multirow{2}{*}{\makecell{Val tokens\\on $P_{\mathrm{val}}$}}} \\
\cmidrule(lr){3-5}
& & \multicolumn{1}{c}{Count} & \multicolumn{1}{c}{Weighted} & \% & & \\
\midrule

\multirow{3}{*}{\makecell[l]{$|V|{=}32k$\\$N{=}100k$}}
& Iter-1 (\(\kappa_p{=}2\))
& 19{,}712 & 9{,}387{,}096 & 1.014
& 986{,}263{,}187 & 370{,}921{,}284 \\

& Iter-2 (\(\kappa_p{=}3\))
& 9{,}785 & 4{,}922{,}576 & 0.531
& 985{,}967{,}502 & 369{,}039{,}323 \\

& Final repricing
& 4{,}605 & 4{,}967{,}839 & 0.536
& 986{,}252{,}419 & 368{,}576{,}235 \\

\midrule

\multirow{3}{*}{\makecell[l]{$|V|{=}64k$\\$N{=}100k$}}
& Iter-1 (\(\kappa_p{=}2\))
& 4{,}214 & 1{,}252{,}383 & 0.135
& 943{,}020{,}279 & 354{,}762{,}017 \\

& Iter-2 (\(\kappa_p{=}3\))
& 438 & 148{,}163 & 0.016
& 942{,}586{,}891 & 354{,}171{,}929 \\

& Final repricing
& 1{,}039 & 480{,}553 & 0.052
& 942{,}556{,}161 & 354{,}108{,}243 \\

\midrule

\multirow{3}{*}{\makecell[l]{$|V|{=}32k$\\$N{=}200k$}}
& Iter-1 (\(\kappa_p{=}2\))
& 62{,}496 & 14{,}968{,}925 & 1.591
& 1{,}024{,}420{,}706 & 369{,}466{,}239 \\

& Iter-2 (\(\kappa_p{=}3\))
& 34{,}314 & 8{,}529{,}593 & 0.907
& 1{,}024{,}042{,}423 & 368{,}120{,}996 \\

& Final repricing
& 19{,}772 & 9{,}225{,}475 & 0.980
& 1{,}025{,}212{,}967 & 367{,}886{,}014 \\

\midrule

\multirow{3}{*}{\makecell[l]{$|V|{=}64k$\\$N{=}200k$}}
& Iter-1 (\(\kappa_p{=}2\))
& 31{,}494 & 4{,}910{,}159 & 0.522
& 978{,}678{,}879 & 352{,}409{,}163 \\

& Iter-2 (\(\kappa_p{=}3\))
& 12{,}902 & 1{,}991{,}109 & 0.212
& 978{,}390{,}997 & 351{,}401{,}677 \\

& Final repricing
& 5{,}500 & 2{,}190{,}078 & 0.233
& 978{,}705{,}696 & 351{,}064{,}086 \\

\bottomrule
\end{tabular*}

\caption{Stage-wise fallbacks for \method{} at $N{=}100k$ and $N{=}200k$ training scopes.
Greedy-L2R token columns score the optimization scope~$P$ and validation holdout~$P_{\mathrm{val}}$, respectively.
The \% column is the weighted fallback mass as a fraction of the full top-$N$ pretoken count mass ($926.2$M at $N{=}100k$, $940.9$M at $N{=}200k$).
}
\label{tab:fallback_diag}
\end{table}

\begin{figure}[t]
\centering
\includegraphics[width=0.98\columnwidth]{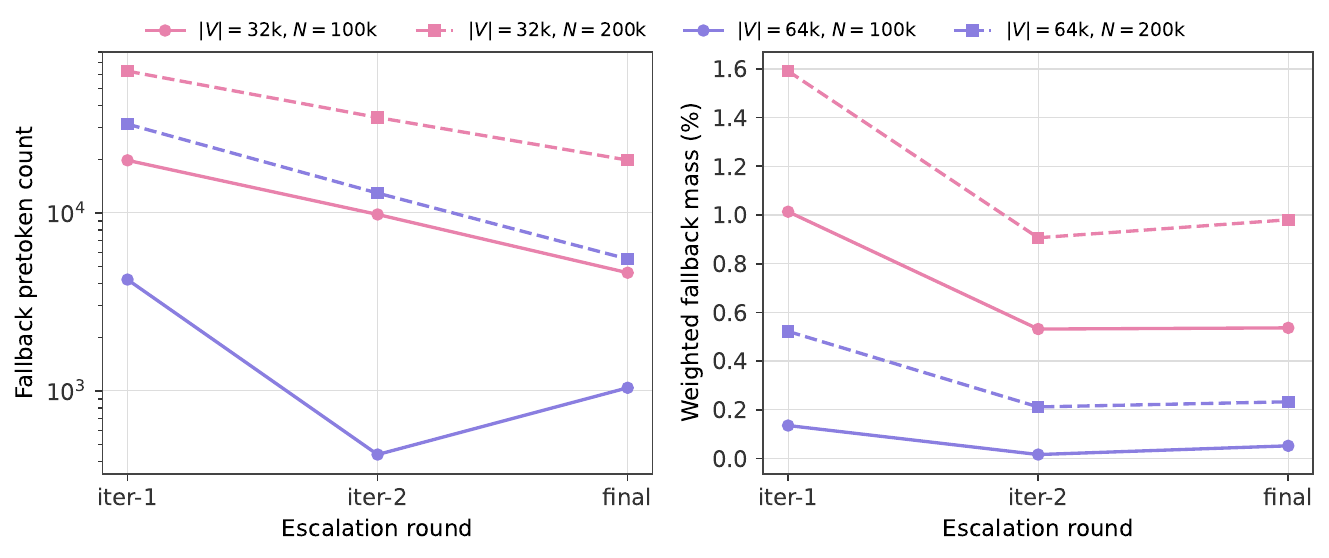}
\caption{Fallback diagnostics per escalation round for all four configurations ($|V|{=}32$k pink, $|V|{=}64$k purple; $N{=}100$k solid, $N{=}200$k dashed). Left: number of pretokens routed to the fallback (log scale). Right: weighted fallback mass as a percentage of the full top-$N$ pretoken count mass (cf.\ \cref{tab:fallback_diag}). The iter-2 dip and slight final-round rise in fallback count reflect the repricing solve (\cref{eq:rev_fb_cost}), while validation scores continue to track the declining fallback mass.}
\label{fig:fallback_mass}
\end{figure}

\section{Token Rank-Frequency on the Validation Holdout}
\label{app:token_zipf}

To test whether \method{}'s compression gains reshape the global vocabulary usage profile or arise from local segmentation choices, we plot log-log token rank--frequency curves after greedy-L2R decoding on the full validation holdout~$P_{\mathrm{val}}$ (\cref{subsec:data_universes_v2}).
Each curve ranks every token emitted on~$P_{\mathrm{val}}$ by its count; BPE-greedy uses a vocabulary trained on the full training holdout, and \method{} curves use the rounded vocabularies from our $N{=}100k$ and $200k$ optimization scopes.

\begin{figure}[t]
\centering
\includegraphics[width=0.95\columnwidth]{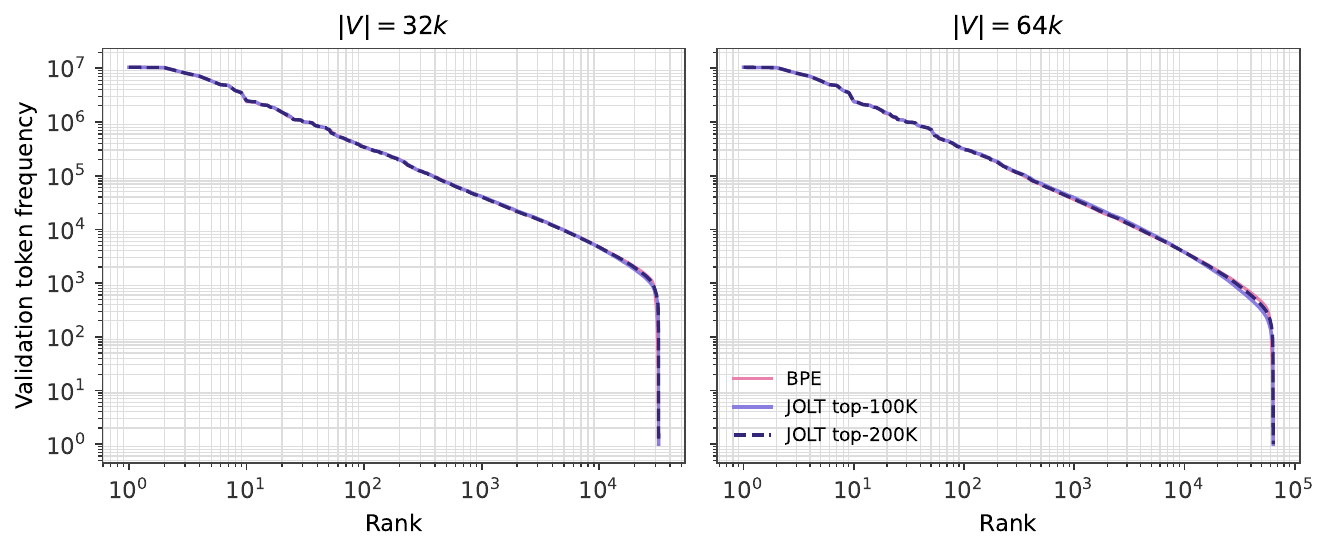}
\caption{Log-log token rank-frequency on the greedy-L2R-decoded validation set~$P_{\mathrm{val}}$ for BPE-greedy and \method{} ($N{=}100k$ and $200k$ optimization scopes), at both vocabulary budgets ($|V|{=}32$k, left; $|V|{=}64$k, right). Series overlap on the high-frequency head; they diverge only in the low-frequency tail at $|V|{=}64$k.}
\label{fig:token_zipf}
\end{figure}

At $|V|{=}32$k (left panel of \cref{fig:token_zipf}), all three series overlap across ranks.
At $|V|{=}64$k (right panel), the high-frequency head remains aligned but the tails separate: $N{=}200k$ tracks BPE-greedy more closely in the rare-token regime than the cheaper $N{=}100k$ scope, consistent with its stronger validation margin in \cref{tab:main_results}.
No series introduces a qualitatively different Zipf shape; the effect is a modest tail adjustment rather than a wholesale redistribution.

\section{Token-Length Distribution: BPE-greedy vs.\ \method{}}
\label{app:token_length_hist}
\begin{figure}[t]
\centering
\includegraphics[width=\linewidth]{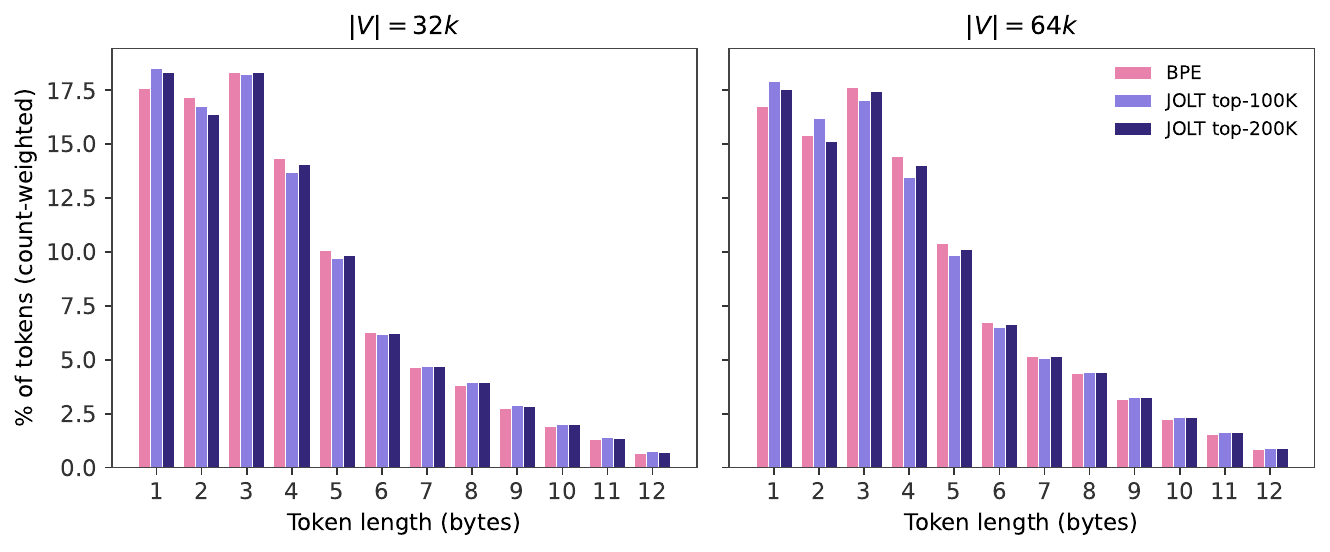}
\caption{Greedy-L2R token-length distribution on~$P_{\mathrm{val}}$ (count-weighted) for BPE-greedy and \method{} at $N{=}100k$ and $200k$, at both vocabulary budgets.}
\label{fig:token_length_hist}
\end{figure}

Beyond aggregate token-count metrics, we compare the \emph{token-length distribution} each vocabulary induces under greedy-L2R decoding on~$P_{\mathrm{val}}$ (\cref{subsec:main_comparison}), because local shifts in emitted token lengths can accompany even small changes in total count.
\cref{fig:token_length_hist} reports count-weighted shares of output tokens by byte length, for both vocabulary budgets and the two full-pipeline scopes ($N{=}100k$ and $N{=}200k$).
BPE-greedy uses a vocabulary trained on the full training holdout; relative to it, \method{} shifts a modest amount of count-weighted mass toward length-1 tokens (by ${\approx}0.9$ percentage points at $|V|{=}32$k and ${\approx}1.2$ at $|V|{=}64$k for the top-$100$K scope), while every other length stays within about $1$ percentage point.
This counter-intuitive increase in single-byte token usage is a consequence of the greedy-consistency constraints: \glr{} can emit a long token at position $i$ only if no longer vocabulary token starts at $i$, so \method{} sometimes retains short (including single-byte) tokens precisely to \emph{suppress} competing prefixes and allow a longer match a few bytes later, a trade that reduces total token count even though it raises the single-byte share.
Enlarging \method{}'s optimization scope from $100k$ to $200k$ pretokens shifts mass slightly away from length-2 tokens (a relative drop of ${\sim}2\%$ at $|V|{=}32$k and ${\sim}7\%$ at $|V|{=}64$k) and into the length-3--4 range, consistent with the LP using its larger budget on a longer pretoken tail.
The shift is modest relative to BPE-greedy, complementing the token-count margins in \cref{tab:main_results}.

\section{Extended-Scope Protocol}
\label{app:extended_scope}

\paragraph{Round-2-only reporting in \cref{tab:lp_round_diag}.}
For $N{=}300$k and $N{=}400$k we report the raw Gurobi iter-2 LP relaxation objective (fallback cost~$3$) and the true greedy-L2R token count on~$P$ from the iter-2 vocabulary.
The LP objective is reported as-is: substituting the true byte-level cost $n_p$ for fallback cost~$3$ in the saved solution $z^\star$ would break the relaxation lower bound---$z^\star$ was optimized under cost~$3$---and can inflate gap-closed above $100\%$.

\paragraph{What the skipped final repricing round would change.}
In iter-2 the fallback cost is $\mathrm{cost}(p,\mathcal{F})=\kappa_p{+}1{=}3$ for every remaining fallback pretoken, regardless of byte length $n_p$ (\cref{eq:rev_fb_cost}).
The full pipeline's final repricing round rebuilds the model with $\mathrm{cost}(p,\mathcal{F})=n_p$, re-solves, and may add whole-word tokens for long rare pretokens.
To see why this matters for validation, consider \texttt{Ġlipopolysaccharide} ($n_p{=}19$ bytes): in iter-2 it is charged only $3$ tokens if it falls back, far below its true cost of $19$ bytes, so Gurobi has no incentive to cover it.
The full pipeline corrects this in the final repricing LP, adding \texttt{Ġlipopolysaccharide}, \texttt{Ġcerebrovascular}, and similar rare but long pretokens as whole tokens, improving validation scores by approximately $0.46$M tokens at $|V|{=}32$k/$N{=}100$k; the round-2-only protocol trades this refinement for the ability to optimize over a much larger pretoken scope.

\paragraph{Sanity check.}
Because the full pipeline remains affordable at $N{=}100$k and $N{=}200$k but not at $N{=}300$k/$400$k, we validate the round-2-only protocol on the smaller scopes before trusting it at scale.
We extract the iter-2 vocabulary and rounded greedy-L2R count on~$P$ from saved checkpoints for the $N{=}100$k and $N{=}200$k runs and compare against their full-pipeline values.
Iter-2 rounded objectives agree with full-pipeline rounded objectives to within $0.03\%$ across all four full-run configurations; validation token counts differ by at most ${\approx}0.5$M tokens.
This close agreement gives us confidence that the round-2-only numbers at $N{=}300$k and $N{=}400$k are comparable to the full-pipeline results.

\section{LP Bound and Rounding Diagnostics (All Training Scopes)}
\label{app:lp_diag}

\Cref{tab:lp_round_diag} reports the LP relaxation objective and the greedy-L2R token count from contribution-based rounding on the \method{} optimization scope~$P$ for all configurations.
For the full-pipeline runs ($N \in \{100\text{k},200\text{k}\}$) the LP objective is the final-round bound with conservative byte-level fallback cost, and is a valid lower bound on achievable \glr{} token count under our formulation.
For the round-2-only runs ($N \in \{300\text{k},400\text{k}\}$) the LP column is the raw Gurobi iter-2 relaxation objective (fallback cost~$3$); the rounded column is the true greedy-L2R count on~$P$ with the iter-2 vocabulary (\cref{app:extended_scope}).
These rows close $75.2$--$86.5\%$ of the BPE-greedy-to-iter-2-LP slack---informative, but not equivalent to the final-round certificate above.

\begin{table}[h]
\centering
\small
\setlength{\tabcolsep}{4pt}
\renewcommand{\arraystretch}{1.12}

\begin{tabular*}{\linewidth}{@{\extracolsep{\fill}}lccccc}
\toprule
Config &
\makecell{LP\\obj.\ (M)} &
\makecell{Rounded\\obj.\ (M)} &
\makecell{Round\\gap \%} &
\makecell{BPE-greedy\\on $P$ (M)} &
\makecell{Gap\\closed \%} \\
\midrule

\multicolumn{6}{l}{\textit{Full pipeline (final repricing LP; valid lower bound)}} \\[2pt]

$|V|{=}32k$, $N{=}100k$
& 984.5 & 986.3 & $+0.176$ & 1{,}001.2 & 89.6 \\

$|V|{=}64k$, $N{=}100k$
& 942.5 & 942.6 & $+0.008$ & 955.1 & 99.4 \\

$|V|{=}32k$, $N{=}200k$
& 1{,}025.1 & 1{,}025.2 & $+0.008$ & 1{,}037.4 & 99.4 \\

$|V|{=}64k$, $N{=}200k$
& 978.0 & 978.7 & $+0.070$ & 988.0 & 93.1 \\

\midrule

\multicolumn{6}{l}{\textit{Round-2-only (raw iter-2 LP$^\dagger$)}} \\[2pt]

$|V|{=}32k$, $N{=}300k$
& 1{,}036.6 & 1{,}040.7 & $+0.398$ & 1{,}053.3 & 75.2 \\

$|V|{=}64k$, $N{=}300k$
& 992.3 & 993.7 & $+0.138$ & 1{,}002.5 & 86.5 \\

$|V|{=}64k$, $N{=}400k$
& 1{,}001.1 & 1{,}002.7 & $+0.163$ & 1{,}011.1 & 83.8 \\

$|V|{=}32k$, $N{=}400k$
& 1{,}046.0 & 1{,}050.6 & $+0.439$ & 1{,}062.8 & 72.5 \\

\bottomrule
\end{tabular*}

\caption{LP bound vs.\ rounded greedy-L2R token count on the \method{} training scope~$P$ for all configurations.
Round gap is $100 \times (\text{rounded} - \text{LP}) / \text{rounded}$.
BPE-greedy uses the same $|V|$ and \glr{} decoding on~$P$ (vocabulary trained on the full training holdout).
Gap closed is $100 \times (\text{BPE-greedy} - \text{rounded}) / (\text{BPE-greedy} - \text{LP})$.
$^\dagger$Round-2-only: iter-2 Gurobi LP objective (fallback cost~$3$); rounded is greedy-L2R on~$P$ with the iter-2 vocabulary. Skipping final repricing leaves fallback costs understated relative to the full pipeline (\cref{subsec:extended_scope,app:extended_scope}), but the raw objective remains a valid relaxation lower bound under the iter-2 formulation.
}
\label{tab:lp_round_diag}
\end{table}

\end{document}